\documentclass[11pt,a4paper]{article}
\usepackage[hyperref]{emnlp2020}
\usepackage{times}
\usepackage{latexsym}
\usepackage{amsmath}
\usepackage{multirow}
\usepackage{booktabs}
\usepackage{graphicx}
\usepackage{subcaption}
\usepackage{bbold}
\usepackage{cleveref}
\usepackage{tabularx}
\usepackage{algorithm}
\usepackage{algpseudocode}
\usepackage{xcolor}
\usepackage{dblfloatfix}

\usepackage{paralist}

\usepackage{microtype}

\aclfinalcopy 
\def\aclpaperid{1208} 

\newcommand{\sarrow}[1][3pt]{\mathrel{%
   \hbox{\rule[\dimexpr\fontdimen22\textfont2-.2pt\relax]{#1}{.4pt}}%
   \mkern-4mu\hbox{\usefont{U}{lasy}{m}{n}\symbol{41}}}}

\title{Lifelong Language Knowledge Distillation}

\author{Yung-Sung Chuang\quad Shang-Yu Su\quad Yun-Nung Chen\\
National Taiwan University, Taipei, Taiwan \\
  \texttt{\{b05901033,f05921117\}@ntu.edu.tw\quad y.v.chen@ieee.org} \\}

\date{}

\begin{document}
\maketitle
\begin{abstract}

It is challenging to perform lifelong language learning (LLL) on a stream of different tasks without any performance degradation comparing to the multi-task counterparts. To address this issue, we present Lifelong Language Knowledge Distillation (L2KD), a simple but efficient method that can be easily applied to existing LLL architectures in order to mitigate the degradation. Specifically, when the LLL model is trained on a new task, we assign a teacher model to first learn the new task, and pass the knowledge to the LLL model via knowledge distillation. Therefore, the LLL model can better adapt to the new task while keeping the previously learned knowledge. Experiments show that the proposed L2KD consistently improves previous state-of-the-art models, and the degradation comparing to multi-task models in LLL tasks is well mitigated for both sequence generation and text classification tasks.\footnote{The source code and data are available at \url{https://github.com/voidism/L2KD}.}

\end{abstract}

\section{Introduction}
\label{sec:intro}

Training a single model to learn a stream of different tasks sequentially usually faces the catastrophic forgetting problem~\citep{mccloskey1989catastrophic}: after learning a new task, the model forgets how to handle the samples from previous tasks. Lifelong learning manages to accumulate the knowledge and retain the performance of previously learned tasks.
It is important especially for real-world natural language processing (NLP) applications, because these applications need to interact with many users from different domains everyday, and the language usage also evolves from time to time. Hence, various NLP tasks have been studied for lifelong learning in the previous work, including sentiment analysis~\citep{chen-etal-2015-lifelong,8101496}, conversational agents~\citep{LLL_chatbot}, word and sentence representation learning~\citep{LLL_word,Liu2019ContinualLF}, text classification, and question answering~\citep{d2019episodic}.

\begin{figure}[t!]
    \centering
    \begin{subfigure}{\columnwidth}
        \includegraphics[width=\linewidth]{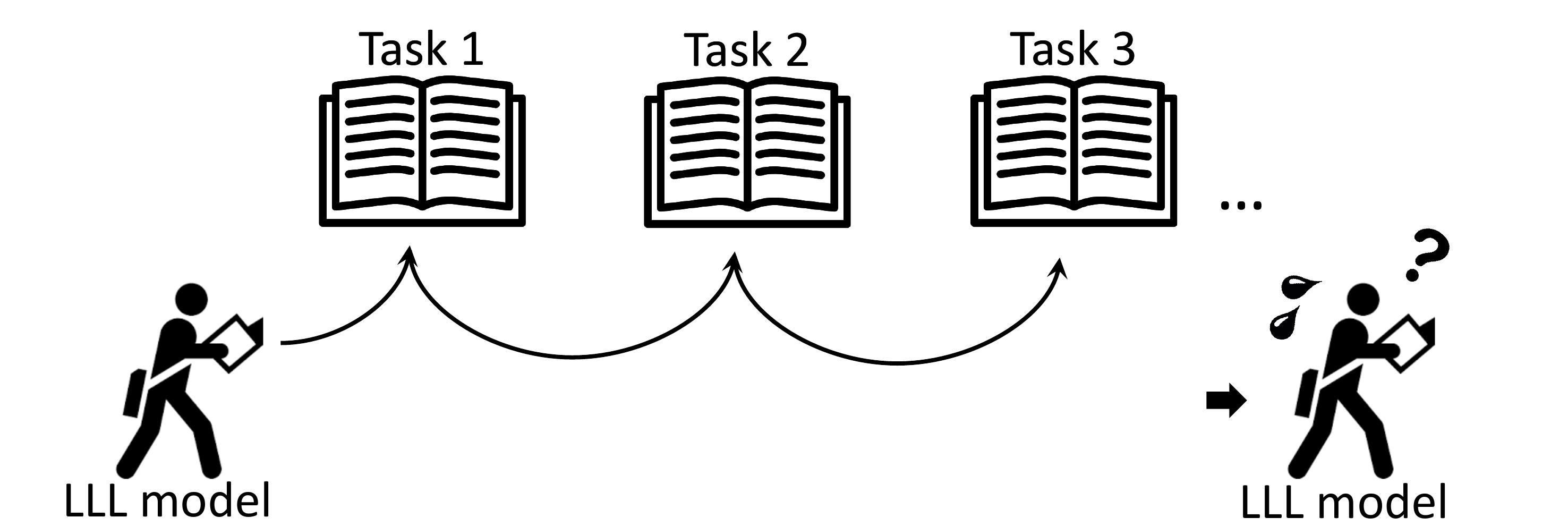}
        \caption{Normal Lifelong Language Learning.}
        \label{fig:anim-lll}
        \end{subfigure}
    \begin{subfigure}{\columnwidth}
        \includegraphics[width=\linewidth]{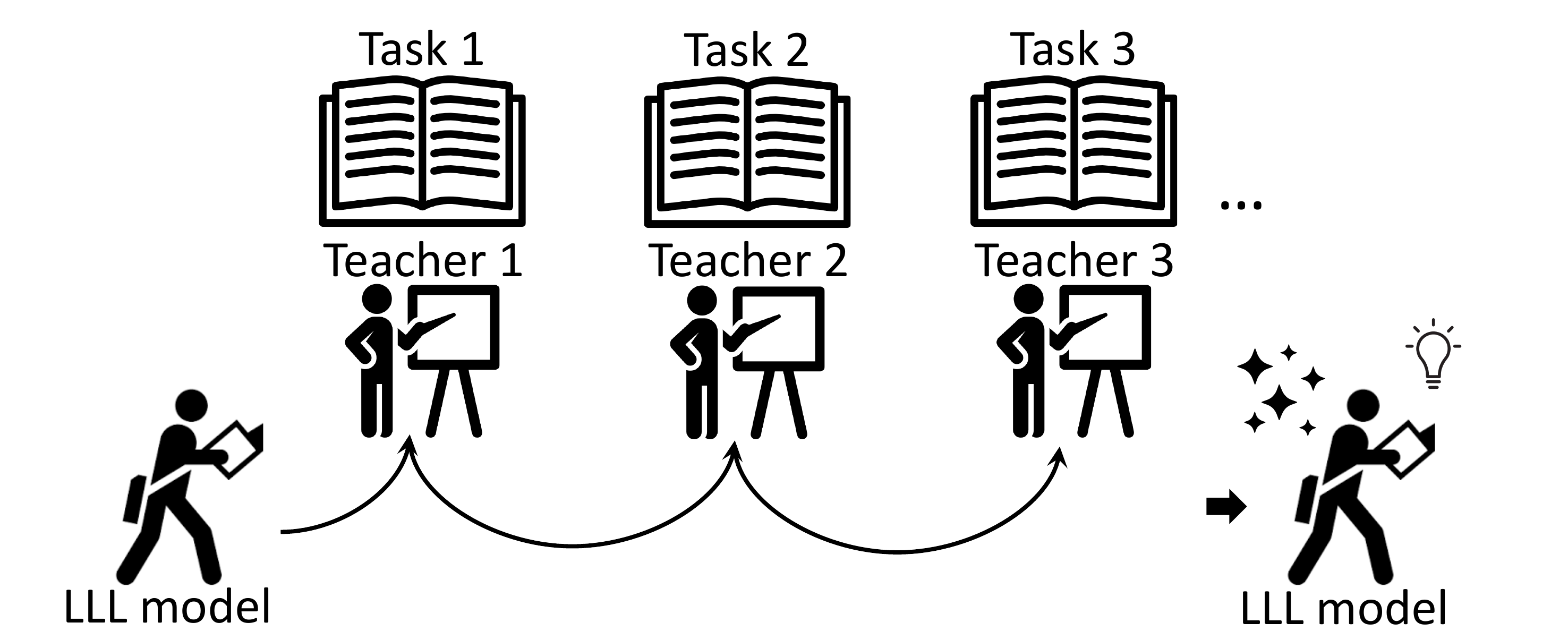}
        \caption{Lifelong Language Knowledge Distillation.}
        \label{fig:anim-l2kd}
    \end{subfigure}
    \caption{The difference between LLL and L2KD.}
    \label{fig:anim-all}
\end{figure}

In recent, LAMOL~\citep{sun2019lamol} improved the performance of LLL by a general framework: 1) it followed the idea about considering many NLP tasks as question answering (QA)~\citep{McCann2018decaNLP} and adapted all tasks into the language modeling (LM) form. In the unified framework, it can perform LLL on many NLP tasks by generating answers based on the contexts and the questions using a single language model, and 2) it outperformed the previous methods by a considerable margin and is only 2\%-3\% worse than the multi-tasking upper bound, which jointly learns all tasks in a mixed dataset.

This paper further improves LLL by introducing Lifelong Language Knowledge Distillation (L2KD), which can be flexibly applied upon the LAMOL architecture or other LLL methods for sequence generation learning. 

The motivation of our work mainly comes from how to efficiently compress the knowledge under a lifelong language learning framework. If the model can learn a new task in an efficient way, the previously learned knowledge may not be affected and thus the problem of catastrophic forgetting can be mitigated.

Inspired by knowledge distillation~\citep{Bucila2006ModelC, Hinton2015DistillingTK, kim2016sequence}, in which a student (smaller) model is trained to imitate the behavior of a teacher (larger) model in order to reach the performance closer to the teacher model, the LLL model in L2KD can be seen as a weak learner that needs to compress knowledge from different tasks into a compact single model. Thus LLL can benefit from the similar procedure of knowledge distillation, although the model size is equal to its teacher model. The similar idea about distilling knowledge from equal-size models has also been studied in 
born-again neural network~\citep{furlanello2018born}, multitask learning~\citep{clark2019bam} and lifelong computer vision learning~\citep{hou2018lifelong}, but never been explored in lifelong language learning research. 

In L2KD, we train a new teacher model when facing a new task, and the LLL model imitates the behavior of its teacher at each training stage, as illustrated in Figure~\ref{fig:anim-all}. This method only needs a little extra time to train a disposable teacher model for each new task, and the teacher model can be discarded when learning the next task; therefore, there is no extra memory or model capacity required for the target LLL model, making the proposed model more memory-efficient for real-world usage.

\section{Proposed Approach}

Before describing how L2KD works, in Section~\ref{ssec:lamol} we briefly introduce the architecture of LAMOL~\citep{sun2019lamol}, which L2KD is built upon. Then we introduce different knowledge distillation strategies in Section~\ref{ssec:KD}, and how to apply them to L2KD in Section~\ref{ssec:L2KD}.

\subsection{LAMOL: Language Modeling for Lifelong Language Learning}
\label{ssec:lamol}

\begin{figure}
    \centering
    \begin{subfigure}{\columnwidth}
        \includegraphics[width=\linewidth]{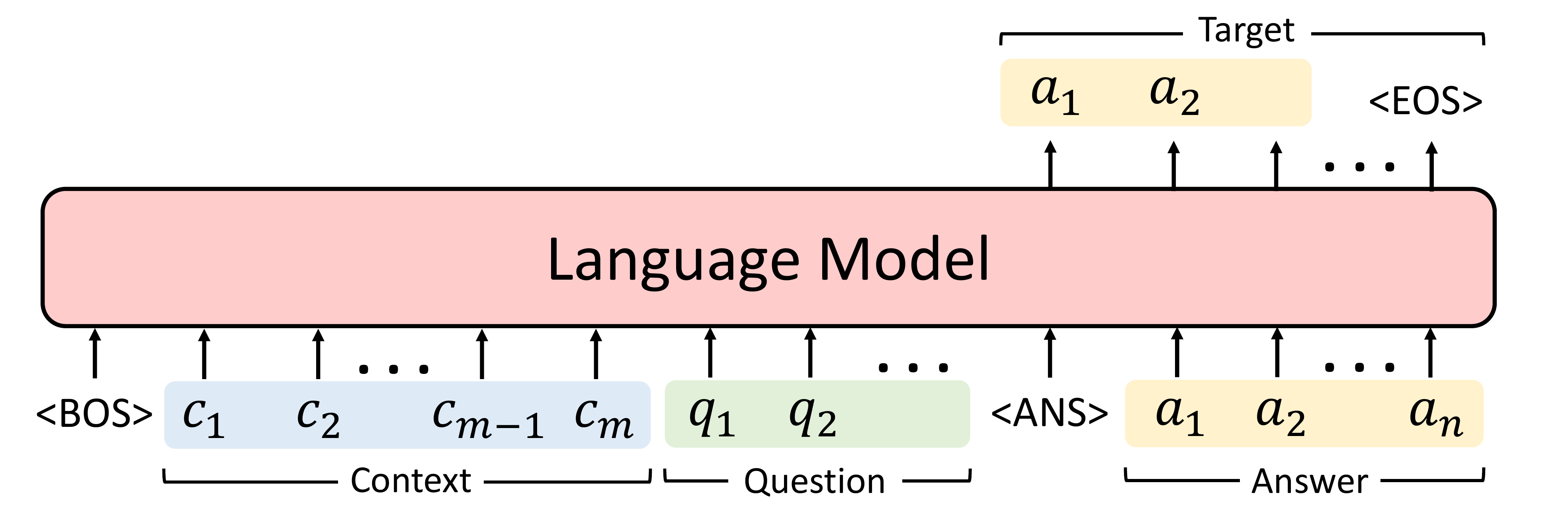}
        \caption{Learning to solve target tasks (QA).}
        \label{fig:lamol-qa}
        \end{subfigure}
    \begin{subfigure}{\columnwidth}
        \includegraphics[width=\linewidth]{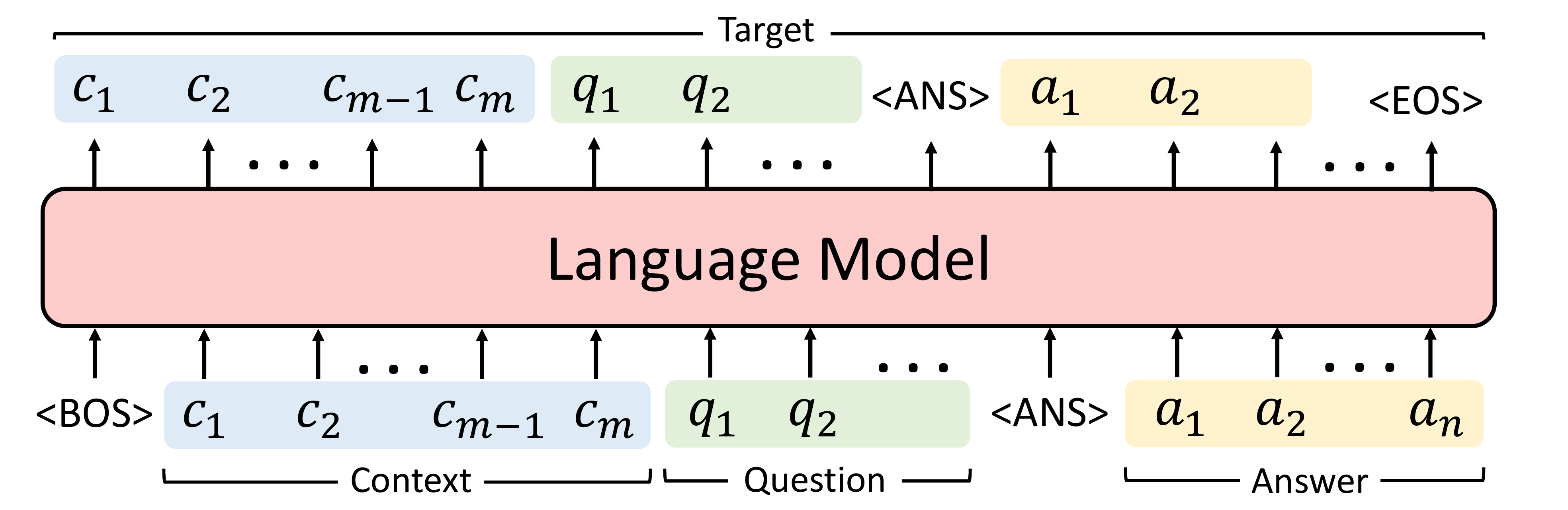}
        \caption{Learning to generate pseudo-data (LM).}
        \label{fig:lamol-lm}
    \end{subfigure}
    \caption{Illustration of learning QA and LM in LAMOL.}
    \label{fig:lamol}
\end{figure}

In the setting of LAMOL, all samples in language datasets have three parts: \emph{context}, \emph{question} and \emph{answer}. We can simply concatenate these three parts into a single sentence and train the model to generate the answer based on the context and question prior to it, as illustrated in Figure~\ref{fig:lamol-qa}.

Besides generating answers for the given questions, the model simultaneously learns to model the whole training sample, as illustrated in Figure~\ref{fig:lamol-lm}. By doing that, when training on the next task, the model can generate training samples for the previous tasks and train on both data from the new task and the generated pseudo-data for the prior tasks. 
Thus the model would forget less when adapting to the new tasks.

LAMOL can outperform previous  regularization-based~\citep{schwarz2018progress, aljundi2018memory} or memory-based~\citep{lopez2017gradient, yogatama2019learning} LLL methods by a large margin. While most of previous methods usually get results slightly better than the finetuning baseline (doing nothing to prevent forgetting), LAMOL already get significant results that are very close to the multitasking upper bound and only 2\%-3\% worse~\citep{sun2019lamol} than it. Thus, in this paper, we focus on how to apply L2KD based on LAMOL.

\subsection{Knowledge Distillation}
\label{ssec:KD}

\paragraph{Language Modeling}
The training objective for normal language modeling is to minimize the negative log-likelihood (NLL) in predicting the next word (hard target):
\begin{equation*}
    \mathcal{L}_\text{NLL}(x; \theta) = \sum_{t=t_0}^{T}{-\log P(x_t \mid x_{<t}; \theta)},
\end{equation*}
where $x_t$ denotes the $t$-th word in the sentence, $x_{<t}$ denotes all words prior to $x_t$, and $\theta$ is the parameters of the language model. 

In knowledge distillation, instead, we minimize the prediction errors between student and teacher models.
The target unit for considering the errors can be done in the word level or the sequence level.

\paragraph{Word-Level (Word-KD)}
We minimize the cross-entropy between the output distributions from student and teacher models when predicting the next word:
\begin{align*}
    \mathcal{L}_{\text{Word-KD}}(x; \theta_S; \theta_T) &=\\ \sum_{t=t_0}^{T}\sum_{k=1}^{|\mathcal V|}-P({\mathcal V}_k \mid &x_{<t}; \theta_{T}) \log P({\mathcal V}_k \mid x_{<t}; \theta_{S}),
\end{align*}
where the input $x_{<t}$ is from the ground truth sequence. $\mathcal V$ denotes the vocabulary set and ${\mathcal V}_k$ is the $k$-th word in $\mathcal V$. $\theta_S$ and $\theta_T$ are parameters of the student model and teacher model respectively.

\paragraph{Sequence-Level (Seq-KD)}
Similar to \citet{kim2016sequence}, we minimize the negative log-likelihood directly on the greedy decode or beam search output sequence $\hat x$ from the teacher model as the hard target, just like normal language modeling:
\begin{equation*}
    \mathcal{L}_{\text{Seq-KD}}(\hat x; \theta_S) = \sum_{t=t_0}^{T}{-\log P(\hat x_t \mid \hat x_{<t}; \theta_S)}.
\end{equation*}
Seq-KD is usually applied for improving weak non-autoregressive translation (NAT) models~\citep{zhou2019understanding} by reducing the multi-modality problem in machine translation datasets~\citep{gu2017non}.

\paragraph{Soft Sequence-Level (Seq-KD$_\text{soft}$)}
We further investigate whether the soft target plus the teacher decoded sequence can help the model more, so we conduct  Seq-KD$_\text{soft}$, in which we perform Word-KD on the greedy decode or beam search outputs from the teacher model. The only difference between Seq-KD$_\text{soft}$ and Word-KD is that the input $x_{<t}$ of Word-KD is now replaced with $\hat x_{<t}$, the output sequence from the teacher model:
\begin{align*}
    \mathcal{L}_{\text{Seq-KD}_\text{soft}}(\hat x; \theta_S; \theta_T) &= \\
    \sum_{t=t_0}^{T}\sum_{k=1}^{|\mathcal V|}-P({\mathcal V}_k \mid &\hat{x}_{<t}; \theta_{T}) \log P({\mathcal V}_k \mid \hat x_{<t}; \theta_{S}).
\end{align*}

Note that no matter what kind of loss we use in knowledge distillation, the teacher model is always fixed. Hence, the optimization process of finding parameters $\theta_S^*$ of the LLL model can be written as follows:
\begin{equation*}
    \theta_S^* = \arg\min_{\theta_S} \mathcal{L}_{\text{KD}}.
\end{equation*}

\subsection{L2KD: Lifelong Language Knowledge Distillation}
\label{ssec:L2KD}
Knowledge distillation can be applied to minimizing both LM and QA loss in LAMOL. Assuming that there is a stream of tasks with datasets $\{D_1, D_2, ...\}$, our LLL model has learned from $D_1$ to $D_{m-1}$ and now was adapted to $D_m$. First we train a teacher model on $D_m$ by minimizing the negative log-likelihood loss both for LM and QA in LAMOL and obtain the model parameters $\theta_T^{m}$.

Now our LLL model (with parameters $\theta_S$) can be trained on $D_m$ by knowledge distillation from the teacher model. Given a training sample $X^m_i = \{x_1, x_2, ..., x_T\} \in D_m$ (including the context, question and answer), we minimize
\begin{align*}
 \mathcal{L_\text{new}}(X^m_i; &\theta_S; \theta_T^m) = \mathcal{L}^{\text{QA}}_\text{new} + \mathcal{L}^{\text{LM}}_\text{new} \\
 \mathcal{L}^{\text{QA}}_\text{new} &= \mathcal{L}_{\text{Word-KD}}(X^m_i; \theta_S; \theta_T^m; t_0=a_1) \\
 \mathcal{L}^{\text{LM}}_\text{new} &= \mathcal{L}_{\text{Word-KD}}(X^m_i; \theta_S; \theta_T^m; t_0=0),
\end{align*}
where $a_1$ denotes the start position of the answer. Here we take Word-KD for illustration, but we can also replace the text in the answer part with the teacher-generated answers, so as to conduct Seq-KD$_\text{soft}$ or Seq-KD.

\begin{algorithm}[t!]
\small
  \caption{L2KD: Lifelong Language Knowledge Distillation}
  \label{alg:L2DK}
  \begin{algorithmic}
    \State \textbf{Input}: current task dataset $D_m$, teacher model with parameters $\theta_T$, knowledge distillation loss function $\mathcal{L}_{\text{KD}}$, pseudo-data sample rate $\gamma$.
    \State \textbf{Output}: LLL model parameters $\theta_S$.
    \State Optimize teacher model on $D_m$ to get parameters $\theta_T$.
    \State Sample $\gamma \cdot |D_m|$ pseudo-data from $\theta_S$ to form $D_\text{prev}$.
    \ForAll{$\text{training samples} \{X^m_i\}^n_{i=1} \in D_m$}  
        \For{$i=1$ {\bfseries to} $n$}
            \State update $\theta_S$ to minimize $\mathcal{L}_{\text{KD}} (X^m_i; \theta_S; \theta_T)$
        \EndFor
        \State Sample $n^\prime = \gamma n$ samples $\{X^\text{prev}_j\}^{n^{\prime}}_{j=1}$ from $D_\text{prev}$
        \For{$j=1$ {\bfseries to} $n^\prime$}
            \State update $\theta_S$ to minimize $\mathcal{L}_{\text{NLL}} (X^\text{prev}_j; \theta_S)$
        \EndFor
    \EndFor
  \end{algorithmic}
\end{algorithm}

Besides training on samples from $D_m$, the LLL model also generates pseudo-data $D_\text{prev}$ for previous tasks.
For samples in $D_\text{prev}$, however, we cannot perform knowledge distillation here, because in our setting the teacher models of previous tasks will be discarded after adapting to the next task. Therefore, given the generated data $X^{\text{prev}}_{i} \in D_\text{prev}$, we only minimize NLL loss here:
\begin{align*}
 \mathcal{L_\text{prev}}(X^\text{prev}_i; &\theta_S) = \mathcal{L}^{\text{QA}}_\text{prev} + \mathcal{L}^{\text{LM}}_\text{prev} \\
 \mathcal{L}^{\text{QA}}_\text{prev} &= \mathcal{L}_{\text{NLL}}(X^\text{prev}_i; \theta_S; t_0=a_1) \\
 \mathcal{L}^{\text{LM}}_\text{prev} &= \mathcal{L}_{\text{NLL}}(X^\text{prev}_i; \theta_S; t_0=0).
\end{align*}

Finally we jointly optimize two loss and obtain the parameters $\theta_S^*$ for the LLL model:
\begin{equation*}
 \theta_S^* = \arg\min_{\theta_S}( \sum_{X^m_i \in D_m}{\mathcal{L_\text{new}}} + \sum_{X^\text{prev}_i \in D_\text{prev}}{\mathcal{L_\text{prev}}})
\end{equation*}
The training procedure is detailed in Algorithm~\ref{alg:L2DK}.

\section{Experimental Setup}
To evaluate the proposed method, we conduct a set of experiments detailed below.

\subsection{Model and Training Details}
\label{ssec:model}
We build our proposed approach based on the implementation of LAMOL\footnote{\url{https://github.com/jojotenya/LAMOL}} to make the results comparable.
We use the same pre-trained small GPT-2~\citep{radford2019language} for all single-task teacher, multitask and LLL models, and train the GPT-2 nine epochs for each dataset.
We use the best setting in LAMOL: using task-specific tokens as begin-of-sentence tokens, and the pseudo-data sample rate is 0.2. During inference, we use greedy decoding to generate sequence. More details can be found in Appendix~\ref{appendix:detail}.

\subsection{Datasets}
\label{ssec:data}

To evaluate the capability of L2KD on diverse sequence generation tasks, we pick the following three tasks from DecaNLP~\citep{McCann2018decaNLP}:
\begin{compactitem}
    \item \textbf{WikiSQL}~\citep{zhong2017seq2sql}: a dataset for developing natural language interfaces for relational databases, in which the model needs to generate structured queries from natural language.
    \item \textbf{CNN/DailyMail}~\citep{see2017get}:
    a text summarization dataset collected from online news articles.
    \item \textbf{MultiWOZ}~\citep{budzianowski2018multiwoz}: a multi-domain wizard-of-oz dataset for task-oriented dialogue modeling, in which the model needs to generate the semantic state sequences based on the given partial dialogues.
\end{compactitem}
Note that we skip machine translation dataset (IWSLT) in DecaNLP here, because GPT-2 does not contain a multilingual vocabulary.
These three datasets focus on \emph{different tasks}, representing the most general case in LLL.

However, in real-world scenarios, it is more common that the LLL model is trained to solve the same task, but in \emph{different domains} that change through time. 
Thus we conduct the experiments on the following natural language generation (NLG) datasets with five different domains:
\begin{compactitem}
    \item \textbf{E2E NLG}~\citep{novikova2017e2e}: a dataset for training end-to-end natural language generation systems in the \textit{restaurant domain}. 
    \item \textbf{RNNLG}~\citep{wen2015semantically}:
    a dataset for NLG in spoken dialogue system application domains. It contains four domains: \textit{San Francisco restaurant search} (rest.), \textit{San Francisco hotel search} (hotel), \textit{Television sale/search} (tv), \textit{Laptop sale/search} (laptop). We use the full dataset for the first three domains and the reduced set for the laptop domain for keeping them balance.
\end{compactitem}

\begin{table}[t!]
\centering
\setlength\tabcolsep{3pt}
\begin{tabular}{llcc}
\toprule
\bf Dataset & \bf Metric & \bf \# Train & \bf \# Test  \\
\midrule
\multicolumn{4}{l}{\it Sequence Generation for Different Tasks}\\
WikiSQL & lfEM & 6,525 & 15,878 \\
CNN/DailyMail & ROUGE & 6,604 & 2,250 \\
MultiWOZ & dsEM & 2,536 & 1,646 \\
\midrule
\multicolumn{4}{l}{\it Sequence Generation for Different Domains}\\
E2E NLG & \multirow{5}{*}{ROUGE} & 6,000 & 2,000 \\
RNNLG (rest.) & \multirow{5}{*}{} & 6,228 & 1,039 \\
RNNLG (hotel) & \multirow{5}{*}{} & 6,446 & 1,075 \\
RNNLG (tv) & \multirow{5}{*}{} & 8,442 & 1,407 \\
RNNLG (laptop) & \multirow{5}{*}{} & 7,944 & 2,649 \\
\midrule
\multicolumn{4}{l}{\it Text Classification for Different Tasks}\\
AGNews & \multirow{5}{*}{Exact Match} & 115,000 & 7,600 \\
Yelp & \multirow{5}{*}{} & 115,000 & 7,600 \\
Amazon & \multirow{5}{*}{} & 115,000 & 7,600 \\
DBPedia & \multirow{5}{*}{} & 115,000 & 7,600 \\
Yahoo & \multirow{5}{*}{} & 115,000 & 7,600 \\
\bottomrule
\end{tabular}
\caption{Dataset sizes and the evaluation metrics.}
\label{tab:datasets}

\end{table}


\begin{table*}[t!]
\centering
\setlength\tabcolsep{3pt}

\begin{tabular}{c|lcccccccccccc}
\toprule
\multicolumn{2}{c}{\textbf{Method}} & \bf WOZ & \bf CNN & \bf SQL & \bf Avg & \bf WOZ & \bf CNN & \bf SQL & \bf Avg & \bf WOZ & \bf CNN & \bf SQL & \bf Avg \\
\midrule
\multicolumn{2}{c}{}& \multicolumn{4}{c}{WOZ $\sarrow$ CNN $\sarrow$ SQL } &\multicolumn{4}{c}{CNN $\sarrow$ SQL $\sarrow$ WOZ } &\multicolumn{4}{c}{SQL $\sarrow$ WOZ $\sarrow$ CNN } \\
\cmidrule(lr){3-6} \cmidrule(lr){7-10}  \cmidrule(lr){11-14}
(a) & Finetune & 0.0 & 26.3 & 64.3 & 30.2 & 84.6 & 6.8 & 2.1 & 31.2 & 0.1 & 26.0 & 0.0 & 8.7 \\
(b) & LAMOL & 67.6 & 27.3 & 62.5 & 52.4 & 83.0 & 27.8 & 60.8 & 57.2 & 76.1 & 26.0 & 55.0 & 52.4 \\
(c) & (b) + Word-KD & \textbf{82.4} & 27.6 & \textbf{65.0} & \textbf{58.3} & \textbf{86.1} & 27.5 & 63.2 & \textbf{59.0} & 79.5 & 26.2 & \textbf{59.6} & 55.1 \\
(d) & (b) + Seq-KD$_\text{soft}$ & 81.0 & 26.9 & 64.7 & 57.5 & 84.1 & 27.6 & \textbf{63.4} & 58.4 & \textbf{81.7} & 25.9 & 58.4 & \textbf{55.3} \\
(e) & (b) + Seq-KD & 76.4 & \textbf{28.0} & 63.7 & 56.1 & 83.0 & \textbf{28.3} & 61.5 & 57.6 & 81.0 & \textbf{27.5} & 57.3 & \textbf{55.3} \\

\midrule
\multicolumn{2}{c}{}&\multicolumn{4}{c}{WOZ $\sarrow$ SQL $\sarrow$ CNN } &\multicolumn{4}{c}{CNN $\sarrow$ WOZ $\sarrow$ SQL } &\multicolumn{4}{c}{SQL $\sarrow$ CNN $\sarrow$ WOZ } \\
\cmidrule(lr){3-6} \cmidrule(lr){7-10}  \cmidrule(lr){11-14}
(a) & Finetune & 0.0 & 25.8 & 0.0 & 8.6 & 3.6 & 24.5 & 64.0 & 30.7 & 85.0 & 7.3 & 0.0 & 30.8 \\
(b) & LAMOL & 76.1 & 26.3 & 59.3 & 53.9 & 79.8 & 27.3 & 64.1 & 57.0 & 84.0 & 27.2 & 58.7 & \textbf{56.6} \\
(c) & (b) + Word-KD & \textbf{81.4} & 26.7 & 59.6 & \textbf{55.9} & 83.5 & 27.8 & \textbf{65.0} & 58.8 & 78.7 & 26.4 & \textbf{59.0} & 54.7 \\
(d) & (b) + Seq-KD$_\text{soft}$ & 80.4 & 26.1 & \textbf{59.9} & 55.5 & \textbf{83.7} & 28.6 & 64.8 & \textbf{59.0} & 84.7 & 26.2 & 58.8 & \textbf{56.6} \\
(e) & (b) + Seq-KD & 77.2 & \textbf{27.0} & 59.5 & 54.5 & 82.8 & \textbf{29.5} & 64.4 & 58.9 & \textbf{84.9} & \textbf{27.8} & 57.3 & \textbf{56.6} \\

\bottomrule
\end{tabular}
\caption{Detailed experimental results on MultiWOZ (WOZ), CNN/DailyMail (CNN), WikiSQL (SQL), with six different lifelong learning orders.}
\label{tab:wozcnnsql_results}
\end{table*}

Although our method is mainly designed for sequence generation tasks, we also use five different text classification datasets to evaluate whether the proposed method also benefits text classification tasks. 
We use the random sampled subsets released by~\citet{sun2019lamol}, each of which has 115,000 training and 7,600 testing instances.
\begin{compactitem}
\item \textbf{AGNews}: News articles, including 4 classes for their topics.
\item \textbf{Yelp}: Customer reviews on Yelp, including 5 classes for their rating scores.
\item \textbf{Amazon}: Customer reviews on Amazon, including 5 classes for their rating scores.
\item \textbf{DBPedia}: Articles on Wikipedia, including 14 classes for their categories.
\item \textbf{Yahoo}: QA pairs on the Yahoo! platform, including 10 classes for their categories.
\end{compactitem}

Due to the limitation of computational resources and the data imbalance, we reduce the big datasets (WikiSQL, CNN/DailyMail, E2E NLG, RNNLG (laptop)) to a smaller size by random sampling. 
The reduced data size and other data statistics in the experiments are detailed in Table~\ref{tab:datasets}.

\section{Results and Discussion}
\label{ssec:results}
We discuss the results for three settings: 1) different sequence generation tasks, 2) same tasks in different domains, and 3) different text classification tasks in order to validate the effectiveness of the proposed approach.

\subsection{Different Sequence Generation Tasks}
\label{ssec:sg}

In the experiments, we perform lifelong learning on the WikiSQL (SQL), CNN/DailyMail (CNN) and MultiWOZ (WOZ) datasets with six different permutation orders, and test the performance at the end of the training streams. 
The detailed results are shown in Table~\ref{tab:wozcnnsql_results}, where the average scores indicate the average of three tasks for overall comparison. Note that the evaluation metrics of these three tasks are all ranging from 0 to 100. The overall results of six orders compared with single-task methods and multitask upper bounds are shown in Table~\ref{tab:average}.

In Table~\ref{tab:wozcnnsql_results}, the first baseline is \textbf{(a) Finetune}, in which we directly train three tasks one after another without preventing catastrophic forgetting.
It is obvious that the Finetune model would forget one or two tasks learned before the final one.
\textbf{(b) LAMOL} is the current state-of-the-art approach that significantly reduce the catastrophic forgetting for comparison.
In the rows (c)-(e), it is shown that applying L2KD upon LAMOL significantly outperforms LAMOL for almost all cases, no matter which knowledge distillation strategy is used: \textbf{(c) Word-KD}, \textbf{(d) Seq-KD$_\text{soft}$}, \textbf{(e) Seq-KD}.
We also observe that among three different knowledge distillation strategies, \textbf{(e) Seq-KD} consistently improves the most on the CNN/DailyMail dataset, which is probably caused by the noisy nature of this summarization dataset.
Therefore, sequence-level knowledge distillation produces a easy-to-learn answer comparing to the original complex answer, so that the LLL model can learn better on it. 

On the other hand, for other two tasks (MultiWOZ, WikiSQL), \textbf{(c) Word-KD} and \textbf{(d) Seq-KD$_\text{soft}$} improve more for most cases.
Because the target sequences of these two tasks are relatively simple, where MultiWOZ focuses on producing semantic state sequences from dialogues, and WikiSQL produces the structured query sequences from the given natural language sentences, the target sequences usually contain the patterns less complex than natural language. So, in these cases, the soft targets may bring more advantages than teacher decoding sequences for the LLL model to learn from.

\begin{table}[t!]
\centering
\setlength\tabcolsep{3pt}
\begin{tabular}{c|lccc|c}
\toprule
\multicolumn{2}{l}{\textit{Non-Lifelong Methods}} & \bf WOZ & \bf CNN & \bf SQL & \bf Avg \\
\midrule
(1)&Single QA & 84.8 & 25.5 & 63.1 & \textbf{57.8} \\
(2)&Single QA+LM & 82.2 & 25.9 & \textbf{63.7} & 57.3 \\
(3)&Multi$_{\text{same}}$ QA & 66.2 & 25.6 & 53.0 & 48.3 \\
(4)&Multi$_{\text{same}}$ QA+LM & 59.0 & 26.3 & 53.6 & 46.3 \\
(5)&Multi$_{\text{long}}$ QA & 82.7 & 26.1 & 61.1 & 56.6 \\
(6)&Multi$_{\text{long}}$ QA+LM & \bf 85.4 & \bf 26.7 & 61.3 & \bf 57.8 \\
\midrule
(7)& (6) + Seq-KD & 84.4 & 27.6 & 61.8 & 58.0 \\
\midrule
\multicolumn{5}{l}{\textit{Lifelong Methods (averaged over six orders)}} \\
\midrule
(a)&Finetune & 28.9 & 19.5 & 21.7 & 23.4 \\
(b)&LAMOL & 77.7 & 27.0 & 60.0 & 54.9 \\
(c)&(b) + Word-KD & 81.9 & 27.0 & \textbf{61.9} & 57.0 \\
(d)&(b) + Seq-KD$_{\text{soft}}$ & \textbf{82.6} & 26.9 & 61.7 & \textbf{57.1} \\
(e)&(b) + Seq-KD & 80.9 & \textbf{28.0} & 60.6 & 56.5 \\
\midrule
\multicolumn{5}{l}{\textit{STD of Lifelong Methods}} \\
\midrule
(f)&Finetune & 43.3 & 9.6 & 32.9 & 28.6 \\
(g)&LAMOL & 6.0 & \textbf{0.7} & 3.2 & 3.3 \\
(h)&(g) + Word-KD & 2.7 & \textbf{0.7} & \textbf{2.8} & 2.1 \\
(i)&(g) + Seq-KD$_{\text{soft}}$ & \textbf{1.8} & 1.0 & 3.0 & \textbf{1.9} \\
(j)&(g) + Seq-KD & 3.4 & 0.9 & 3.1 & 2.5 \\

\bottomrule
\end{tabular}
\caption{Averaged results at the final task of the lifelong learning procedures over six orders, comparing to single task and multitask upper bound. The bold numbers are the best in the group.}
\label{tab:average}
\end{table}

In Table~\ref{tab:average}, the overall performance (averaged over six permutation orders) is compared with single-task methods and multi-task upper bounds.
There are two training methods here: optimizing QA loss only (in rows (1)(3)(5)) or optimizing both QA and LM loss (in rows (2)(4)(6)), as illustrated in Figure~\ref{fig:lamol}.  For multi-task models, we find that the same training steps (9 epochs on the mixed set) may not lead the models to converge (in row (3)(4)), so we additionally train multi-task models for three times longer (27 epochs on the mixed set) in rows (5)(6).

The second part of Table~\ref{tab:average} shows the average performance in lifelong learning of six permutation orders.
It is clear that L2KD significantly improves the average score from 54.9 in \textbf{(b) LAMOL} to 57.1 in \textbf{(d) Seq-KD$_\text{soft}$}. 
The performance of Seq-KD$_\text{soft}$ is only 0.7\% worse than the multi-task upper bound, 57.8 in \textbf{(6) Multi$_\text{long}$ QA+LM}. Hence, the results show that L2KD can bridge the gap between lifelong learning and multi-task learning.

Note that we can also apply similar distillation strategy on multitask learning to obtain a stronger upper bound, which might be a more fair comparison. Thus, we add Seq-KD to \textbf{(6) Multi$_\text{long}$ QA+LM} by making the model learn from five single-task teachers and the results are shown in row (7). We observe that the improvement on multitask learning is only 0.2\%, while L2KD can improve LAMOL by 2.2\%. This result indicates that the benefits brought by knowledge distillation may be saturated for multitask learning, but is not saturated for L2KD. The gap between lifelong learning and multi-task learning is still reduced even if we apply similar strategy on both of the models.

The third part of Table~\ref{tab:average} shows the standard deviations of six permutation orders. 
As mentioned in \citet{sun2019lamol}, if an algorithm has smaller standard deviation over different training orders, it means that the algorithm is more robust and not susceptible to learning orders.
It can be found that the average standard deviation of LAMOL is reduced from 3.3 to 1.9 with \textbf{Seq-KD$_\text{soft}$}.
Therefore, both soft target training and teacher decode sequence can stabilize the training process of LLL and make it more order-agnostic.

\begin{figure}[t!]
    \centering
    \begin{subfigure}{\columnwidth}
        \includegraphics[width=\linewidth]{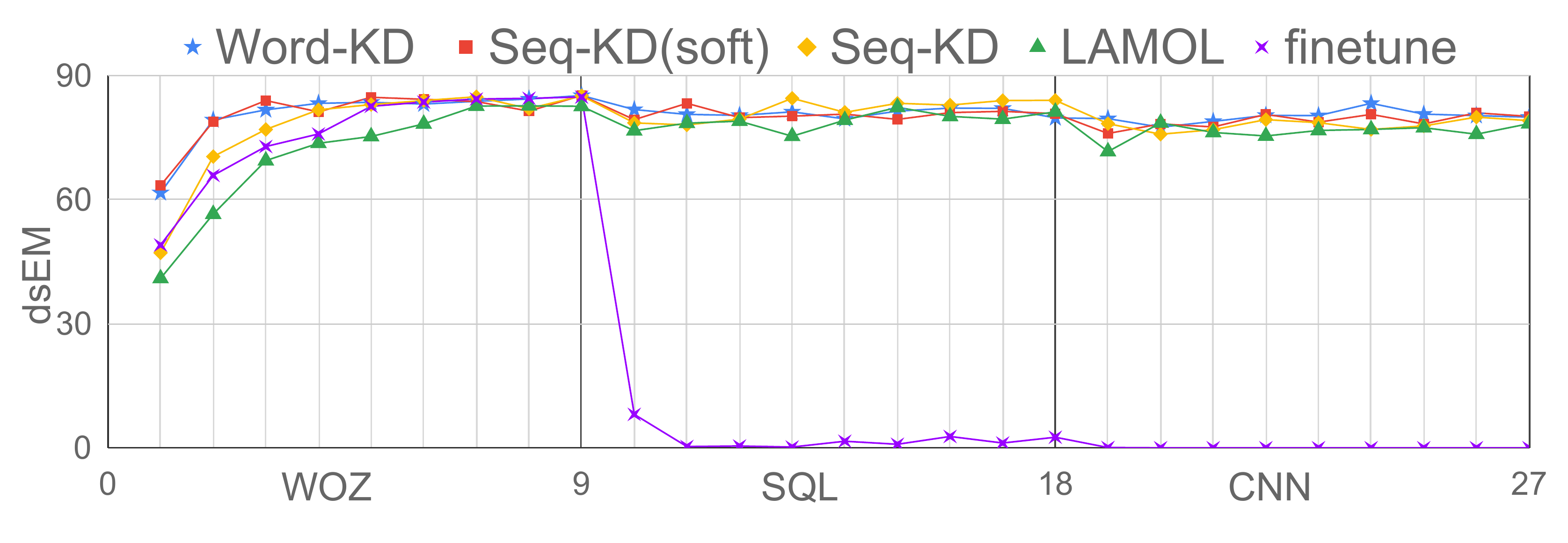}
        \caption{Learning curve of WOZ.}
        \label{fig:curve_woz}
        \end{subfigure}
    \begin{subfigure}{\columnwidth}
        \includegraphics[width=\linewidth]{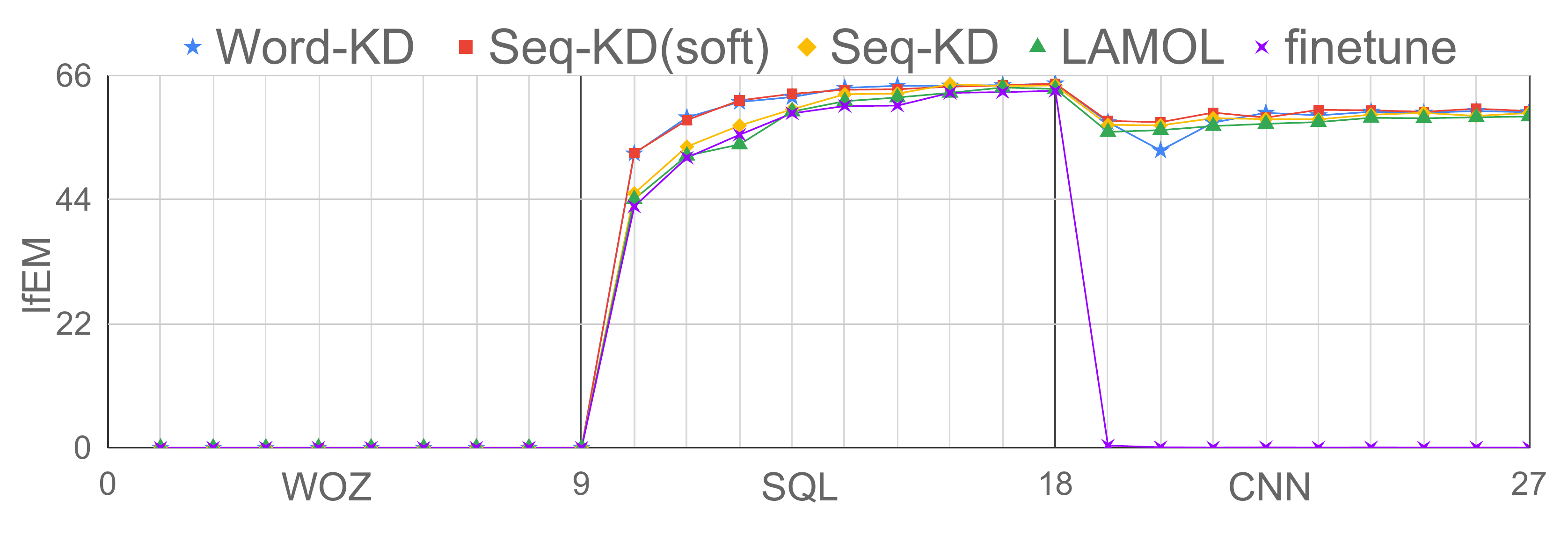}
        \caption{Learning curve of SQL.}
        \label{fig:curve_sql}
    \end{subfigure}
    \begin{subfigure}{\columnwidth}
        \includegraphics[width=\linewidth]{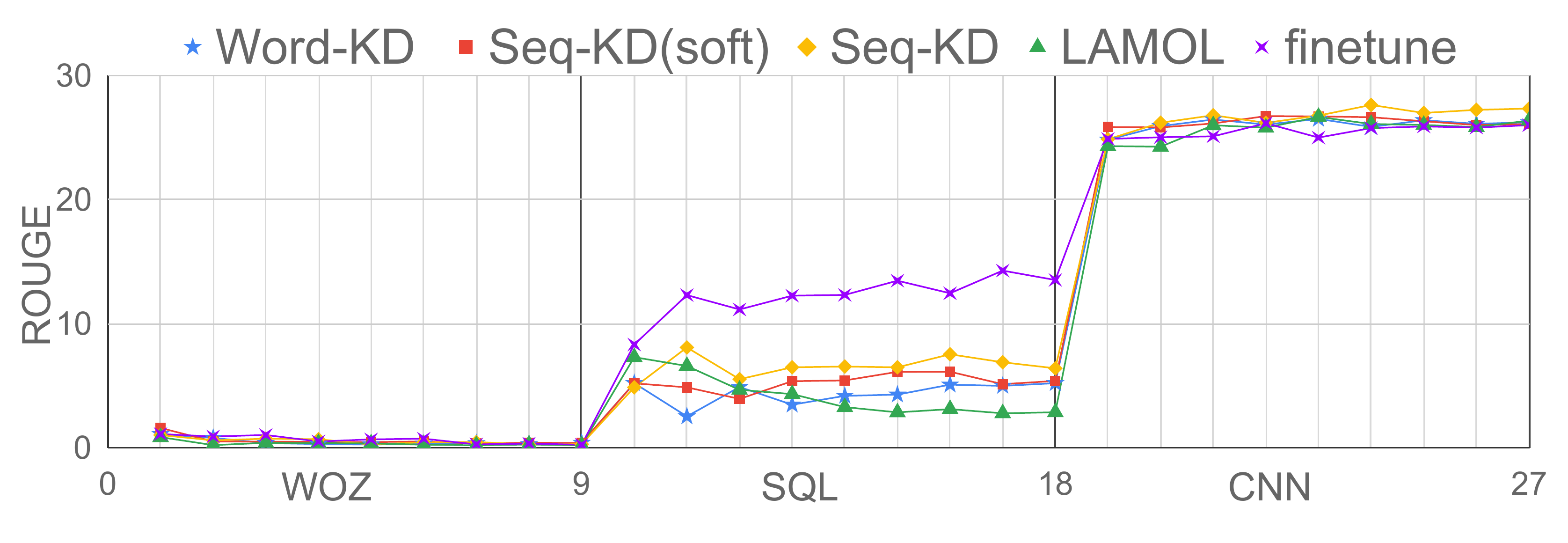}
        \caption{Learning curve of CNN.}
        \label{fig:curve_cnn}
    \end{subfigure}
    \caption{The learning curves of different LLL methods in the order of WOZ $\rightarrow$ SQL $\rightarrow$ CNN.}
    \label{fig:curve}
\end{figure}

To further analyze the performance change when training on different tasks, we plot the testing results during whole lifelong learning stages with the order of WOZ (1-9 epoch) $\rightarrow$ SQL (10-18 epoch) $\rightarrow$ CNN (19-27 epoch) in Figure~\ref{fig:curve}.
In Figure~\ref{fig:curve_woz}, the performance of WOZ for all methods is illustrated. The finetune baseline (purple line) significantly degrades when moving to the next task (SQL) in the second training stage, while other methods can keep the performance.
We observe that applying soft-target Word-KD (blue) or Seq-KD$_\text{soft}$ (red) can increase the scores  faster than hard-target Seq-KD (yellow) and LAMOL baseline (green) at the initial epochs, indicating the effectiveness of the proposed L2KD.
In terms of other two tasks, all distillation methods (Word-KD, Seq-KD$_\text{soft}$, Seq-KD) are capable of maintaining the performance of WOZ slightly better than LAMOL, and finally converge to better points in the third training stage. 
A similar trend can be observed in Figure~\ref{fig:curve_sql}, where soft-target Word-KD and Seq-KD$_\text{soft}$ rise faster in the second training stage and finally drop less than original LAMOL in the third training stage, demonstrating the great property of our proposed methods as LLL models.

In Figure~\ref{fig:curve_cnn}, in the third stage, the yellow line (Seq-KD) converges to a better point than  all other methods, because it reduces the complexity of the noisy summarization dataset. 
However, although Seq-KD$_\text{soft}$ also reduces the complexity, it does not achieve the same performance as Seq-KD.
The probable reason is that the teacher decoding sentences may be easy enough for the LLL model to learn from, and the soft target here 
makes the model not completely converge on these easy sentences.

\begin{table}[t!]
\centering
\small
\setlength\tabcolsep{3pt}
\begin{tabular}{lccccc|c}
\toprule
\textbf{Method}  & \textbf{e2e} & \textbf{rest} & \textbf{hotel} & \textbf{tv} & \textbf{laptop} & \textbf{Avg} \\
\midrule
Single$_\text{(QA)}$ & 48.8 & 64.0 & 65.4 & 70.8 & 73.0 & 64.4 \\
Single$_\text{(QA+LM)}$ & 48.8 & 64.2 & 65.5 & 71.0 & 72.8 & 64.5 \\
Multi$_\text{(QA)}$ & 49.2 & \bf 65.6 & \bf 67.2 & 72.7 & \bf 74.8 & \bf 65.9 \\
Multi$_\text{(QA+LM)}$ & \bf 49.5 & 65.2 & 66.7 & \bf 73.4 & 74.6 & \bf 65.9 \\

\midrule
\multicolumn{7}{l}{\textit{Left-to-right} (e2e $\sarrow$ rest $\sarrow$ hotel $\sarrow$ tv $\sarrow$ laptop)}\\
\midrule
LAMOL & \textbf{50.1} & 58.7 & 61.5 & 73.7 & 72.0 & 63.2 \\
+ Word-KD & 44.9 & 60.0 & 62.8 & \bf 76.7 & 73.3 & 63.5 \\
+ Seq-KD$_\text{soft}$ & 46.9 & 58.4 & 63.2 & 76.4 & 73.6 & 63.7 \\
+ Seq-KD & 48.6 & \textbf{62.2} & \textbf{66.4} & 74.7 & \textbf{75.5} & \textbf{65.5} \\

\midrule
\multicolumn{7}{l}{\textit{Right-to-left} (laptop $\sarrow$ tv $\sarrow$ hotel $\sarrow$ rest $\sarrow$ e2e)}\\
\midrule
LAMOL & \textbf{49.8} & 65.0 & 65.9 & 75.8 & 77.0 & 66.7 \\
+ Word-KD & 49.3 & \textbf{67.6} & \textbf{68.7} & 76.8 & 77.7 & \textbf{68.0} \\
+ Seq-KD$_\text{soft}$ & 49.4 & 66.6 & 68.0 & 76.7 & 77.4 & 67.6 \\
+ Seq-KD & 49.7 & 65.9 & 66.7 & \textbf{77.4} & \textbf{78.8} & 67.7 \\

\bottomrule
\end{tabular}
\caption{Experimental results on NLG datasets from different domains.}
\label{tab:nlg}
\end{table}

\subsection{Same Task in Different Domains}
\label{ssec:nlg}

We perform L2KD on the same NLG task with five different domains: \textit{restaurant} from E2ENLG, \textit{restaurant/hotel/tv/laptop} from RNNLG. Note that although both E2ENLG and RNNLG has the \textit{restaurant} domain, their input formats and label types are totally different.
The results are shown in Table~\ref{tab:nlg}, where we only show two orders in the experiments: from the hardest task to the easiest one (left-to-right) and its reverse order (right-to-left)\footnote{The shown results are representative among all others.}. 
The results show that L2KD outperforms original LAMOL for most cases and improves the averaged ROUGE score by nearly 2 points.

\begin{figure}[t]
    \centering
    \begin{subfigure}{\columnwidth}
        \includegraphics[width=\linewidth]{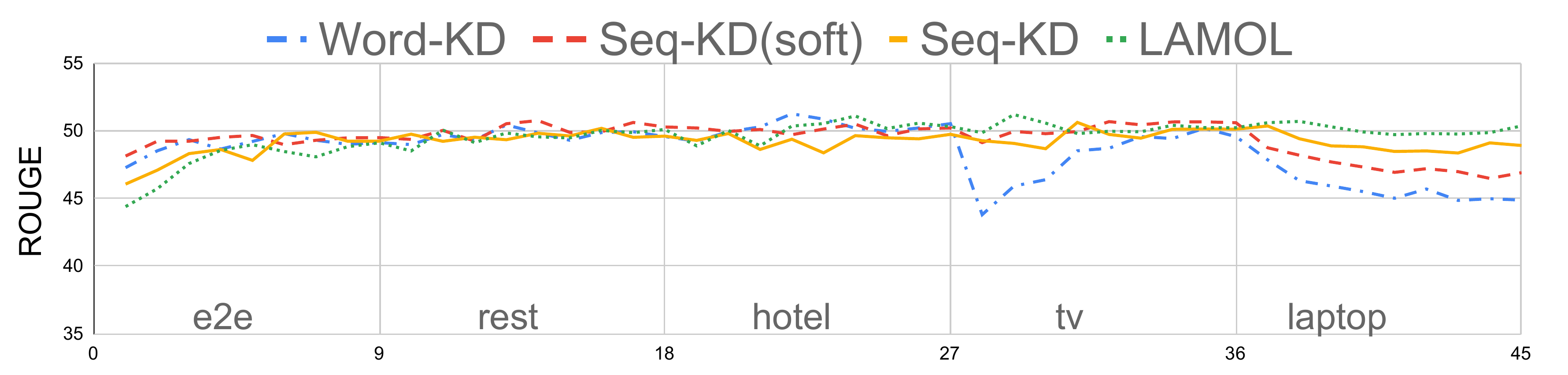}
        \caption{Learning curve of E2ENLG.}
        \label{fig:curve_nlg1}
        \end{subfigure}
    \begin{subfigure}{\columnwidth}
        \includegraphics[width=\linewidth]{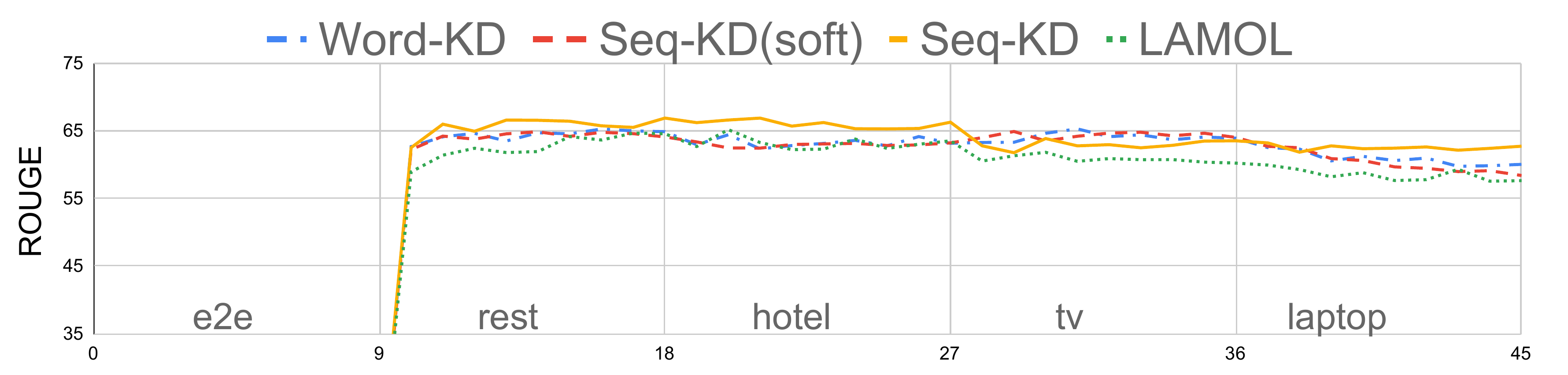}
        \caption{Learning curve of RNNLG (restaurant).}
        \label{fig:curve_nlg2}
    \end{subfigure}
    \begin{subfigure}{\columnwidth}
        \includegraphics[width=\linewidth]{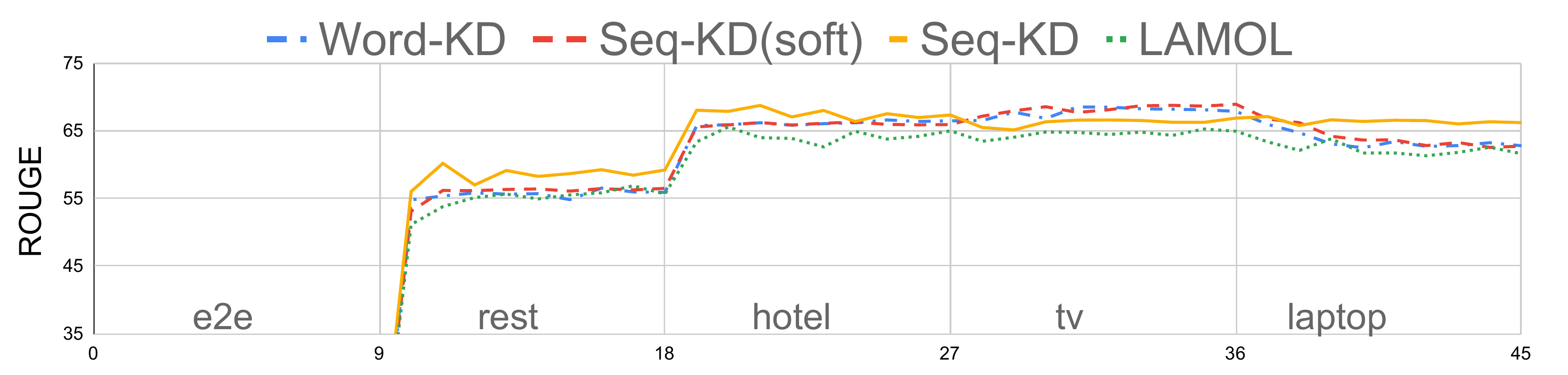}
        \caption{Learning curve of RNNLG (hotel).}
        \label{fig:curve_nlg3}
    \end{subfigure}
    \begin{subfigure}{\columnwidth}
        \includegraphics[width=\linewidth]{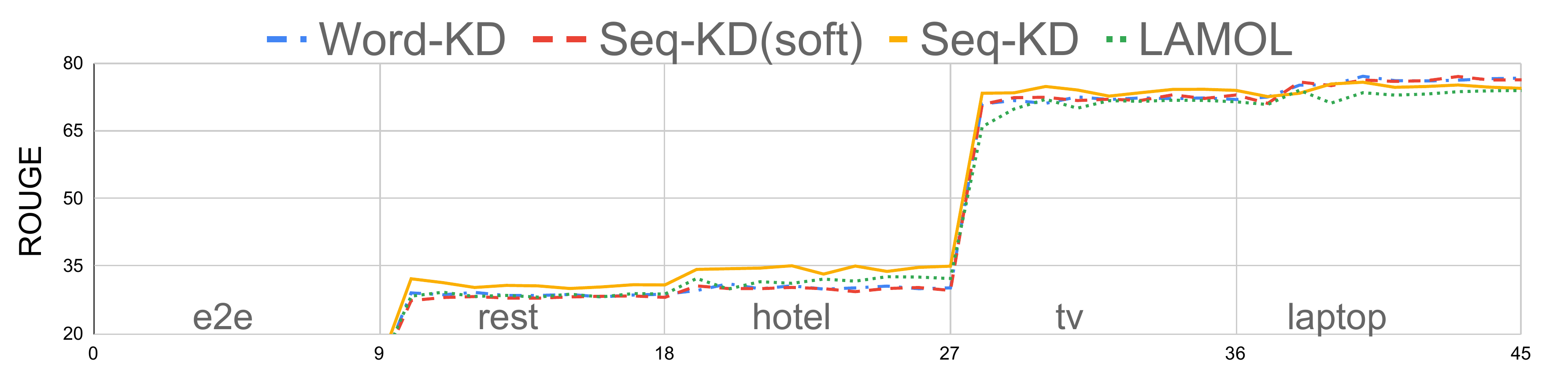}
        \caption{Learning curve of RNNLG (tv).}
        \label{fig:curve_nlg4}
    \end{subfigure}
    \begin{subfigure}{\columnwidth}
        \includegraphics[width=\linewidth]{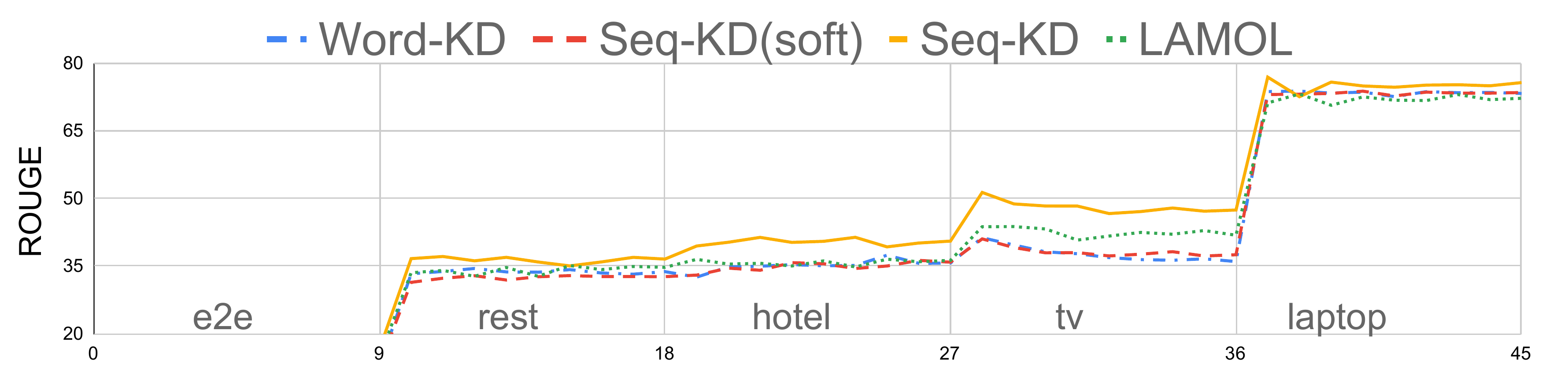}
        \caption{Learning curve of RNNLG (laptop).}
        \label{fig:curve_nlg5}
    \end{subfigure}
    \caption{The learning curves on NLG tasks with the hardest-to-easiest (left-to-right) order.}
    \label{fig:curve_nlg}
\end{figure}

We find that different training orders bring slightly different results. 
In the right-to-left order, the baseline LAMOL can easily outperform multi-task models due to its easiest-to-hardest order, which helps the model to better transfer the knowledge gradually in these NLG tasks, similar to curriculum learning~\citep{bengio2009curriculum}. 
Therefore, it does not mean that lifelong learning can beat multi-task model in all the experiment.

We plot the learning curves of these five tasks in left-to-right order in Figure~\ref{fig:curve_nlg} for further analysis.
Except for E2ENLG, the Seq-KD in yellow lines usually gain more performance at the end of the training stream. Also, we observe that when forward transfer exists, Seq-KD usually benefits more. 
For example, in Figure.~\ref{fig:curve_nlg3}, when training on RNNLG (restaurant) in 10-18th epochs, the ROUGE score on RNNLG (hotel) has risen even before the model first sees RNNLG (hotel) data at the 19th epoch, indicating that the forward transfer exists in this order.
The rising trend is more obvious in Seq-KD (yellow), and the same trend can also be obversed in Figure~\ref{fig:curve_nlg4} and~\ref{fig:curve_nlg5}.

\subsection{Text Classification Tasks}
\label{ssec:tc}

Although our method is mainly designed for sequence generation tasks, we investigate whether this idea also benefits text classification (TC) tasks.
Thus we perform L2KD on five TC tasks, where the answers are always very short sequences representing the class labels of the given documents, such as \texttt{World, Sports, Business, or Sci/Tech} in the AGNews dataset.
Hence, generating such short answers is not complex for the proposed model, and the performance mainly reflects the text understanding performance instead of the generating capability.

We also conduct the experiments from the hardest task to the easiest task, and its reverse order shown in Table~\ref{tab:tcfull}. 
To our surprise, L2KD also improves LAMOL to get better results on TC tasks. The results of these two orders are only 0.1\% worse than the multi-task upper bounds. The Word-KD improves the most on these TC tasks in most cases, and the improvements are more obvious especially for the earlier learned tasks. The details of the learning curves in TC tasks are also shown in Appendix~\ref{appendix:curve_tc} for reference. 

\begin{table}[t!]
\centering
\small
\setlength\tabcolsep{3pt}
\begin{tabular}{lccccc|c}
\toprule
\textbf{Method} & amazon & yelp & yahoo & ag & dbpedia & Avg \\
\midrule

Single$_\text{(QA)}$ & 55.9 & 63.3 & 70.6 & 93.6 & 99.0 & 76.5 \\
Single$_\text{(QA+LM)}$ & 56.9 & \textbf{64.5} & 70.1 & 93.7 & \textbf{99.1} & 76.9 \\
Multi$_\text{(QA)}$ & 56.6 & 63.3 & 69.2 & 93.7 & 99.0 & 76.4 \\
Multi$_\text{(QA+LM)}$ & \textbf{57.8} & 64.4 & \textbf{70.9} & \textbf{94.0} & \textbf{99.1} & \textbf{77.2} \\

\midrule
\multicolumn{7}{l}{\textit{Left-to-right} (amazon $\sarrow$ yelp $\sarrow$ yahoo $\sarrow$ ag $\sarrow$ dbpedia)}\\
\midrule
LAMOL & 52.7 & 61.6 & 70.3 & 93.6 & 99.1 & 75.5 \\
+ Word-KD & \textbf{57.5} & \textbf{63.6} & \textbf{71.3} & \textbf{93.9} & \textbf{99.2} & \textbf{77.1} \\
+ Seq-KD$_\text{soft}$ & 55.7 & 62.0 & \textbf{71.3} & \textbf{93.9} & \textbf{99.2} & 76.4 \\
+ Seq-KD & 56.8 & 62.3 & 71.1 & 93.4 & 99.1 & 76.6 \\

\midrule
\multicolumn{7}{l}{\textit{Right-to-left} (dbpedia $\sarrow$ ag $\sarrow$ yahoo $\sarrow$ yelp $\sarrow$ amazon)}\\
\midrule
LAMOL & 57.9 & 63.5 & 70.7 & 91.7 & 98.3 & 76.4 \\
+ Word-KD & 57.0 & 64.1 & \textbf{73.2} & \textbf{92.7} & \textbf{98.8} & \textbf{77.1} \\
+ Seq-KD$_\text{soft}$ & 57.0 & 64.1 & 71.9 & 92.4 & \textbf{98.8} & 76.8 \\
+ Seq-KD & \textbf{58.4} & \textbf{64.4} & 71.7 & 91.5 & \textbf{98.8} & 76.9 \\
\bottomrule
\end{tabular}
\caption{Experimental results on five text classification datasets.}
\label{tab:tcfull}
\end{table}

Because the answers in TC tasks are not as complex as other sequence generation tasks, we investigate where the improvement mainly comes from during the distillation process.
Therefore, we split each testing set into two groups: (A) \emph{questions correctly answered by the teacher model}; (B) \emph{questions incorrectly answered by the teacher model}. We suspect that the LLL model trained by L2KD may totally copy the behavior from the teacher models and get improvement mainly from the group (A), while it fails to answer the questions in the group (B). To figure it out, we compute the accuracy of each LLL model (in left-to-right experiment) for the groups (A) and (B) respectively, and the difference between original LAMOL and three distillation strategies on five tasks. The averaged results are shown in Table~\ref{tab:tcanalysis_avg}, and the more detailed results for each task can be found in Appendix~\ref{appendix:tcanalysis}.
Surprisingly, applying L2KD does not largely degrade the accuracy in the group (B) comparing to the original LAMOL, and even improves for Word-KD, showing that the LLL model does not fully copy the behavior from its teacher models. On the other hand, the total improvement indeed mainly comes from the group (A), and Word-KD also can improve the most. The double improvement both on group (A) and (B) for Word-KD indicates that on these TC tasks, the LLL model trained by Word-KD can better reach the balance between the \emph{teacher knowledge} and the \emph{transfer ability}. Therefore, it can get the advantages from the teacher knowledge while avoid some false knowledge taught from its teacher by integrating the knowledge from other tasks.

\begin{table}[t!]
\centering
\small
\setlength\tabcolsep{3pt}
\begin{tabular}{lccc}
\toprule
& \bf Acc & \bf Acc in (A) & \bf Acc in (B)\\
\midrule
Teacher & 76.73 & 100.00 & 0.00 \\
\midrule
LAMOL & 75.48 & 88.15 & 33.69 \\
\midrule
+ Word-KD & 77.11 & 90.26 ({\color{green} +2.11}) & 33.75 ({\color{green} +0.06}) \\
+ Seq-KD$_\text{soft}$ & 76.42 & 89.42 ({\color{green} +1.27}) & 33.52 ({\color{red} -0.17}) \\
+ Seq-KD & 76.56 & 89.56 ({\color{green} +1.41}) & 33.69 ({\color{black} +0.00}) \\
\bottomrule
\end{tabular}
\caption{The accuracy in the group (A) and (B) averaged over five classification datasets. The teacher scores are from five single-task models.}
\label{tab:tcanalysis_avg}
\end{table}

\section{Related Work}
\label{sec:relatedwork}

Knowledge distillation has been introduced to the field of lifelong learning; for example, Learning without Forgetting (LwF)~\citep{li2017learning}, Generative Replay with Distillation (DGR+distill), Replay-through-Feedback (RtF)~\citep{van2018generative}, and Lifelong GAN~\citep{zhai2019lifelong}, a lot of prior studies have also used knowledge distillation in lifelong learning, but all in computer vision tasks.
Different from the prior work, this paper is the first attempt that adopts knowledge distillation for NLP tasks in the lifelong learning framework.
Moreover, the prior work used the old model as a teacher to help the current model retain the knowledge about previous tasks. In contrast, our method trains a new teacher model on the incoming new task.
Thus, these two directions of applying knowledge distillation are complementary to each other, showing the potential of applying the proposed method to the fields other than NLP.

\section{Conclusion}
\label{sec:conclusion}

This paper presents Lifelong Language Knowledge Distillation (L2KD), a simple method that effectively help lifelong language learning models to maintain good performance comparable to its multi-task upper bounds. 
The experiments show the consistent improvement achieved by L2KD for three different settings, indicating the effectiveness of the proposed method to train robust LLL models.
In addition, the proposed approach only requires a little extra time for training the teacher without extra memory or capacity needed, showing the potential of being applied to the practical scenarios.

\section*{Acknowledgments}

We thank reviewers for their insightful comments.
We are grateful to Cheng-Hao Ho and Fan-Keng Sun for their source code of LAMOL and valuable suggestions.
This work was financially supported from the Young Scholar Fellowship Program by Ministry of Science and Technology (MOST) in Taiwan, under Grant 109-2636-E-002-026.

\bibliography{emnlp2020}
\bibliographystyle{acl_natbib}

\clearpage
\newpage

\appendix

\section{Training Details}
\label{appendix:detail}
We use a single NVIDIA TESLA V100 (32G) for each experiment. The average runtime of experiments in Section~\ref{ssec:sg} and \ref{ssec:nlg} are 3-8 hours. The experiments in Section~\ref{ssec:tc} need about 3 days for a single experiment.

We did not conduct hyperparameter search, but follow all best settings in the official implementation of LAMOL \footnote{\url{https://github.com/jojotenya/LAMOL}} to keep the results comparable. The main hyperparameters are listed in Table~\ref{tab:hyper}. More details can be found in our released code.

\section{Learning Curves for Text Classification Tasks}
\label{appendix:curve_tc}

\begin{figure}[h]
    \centering
    \begin{subfigure}{\columnwidth}
        \includegraphics[width=\linewidth]{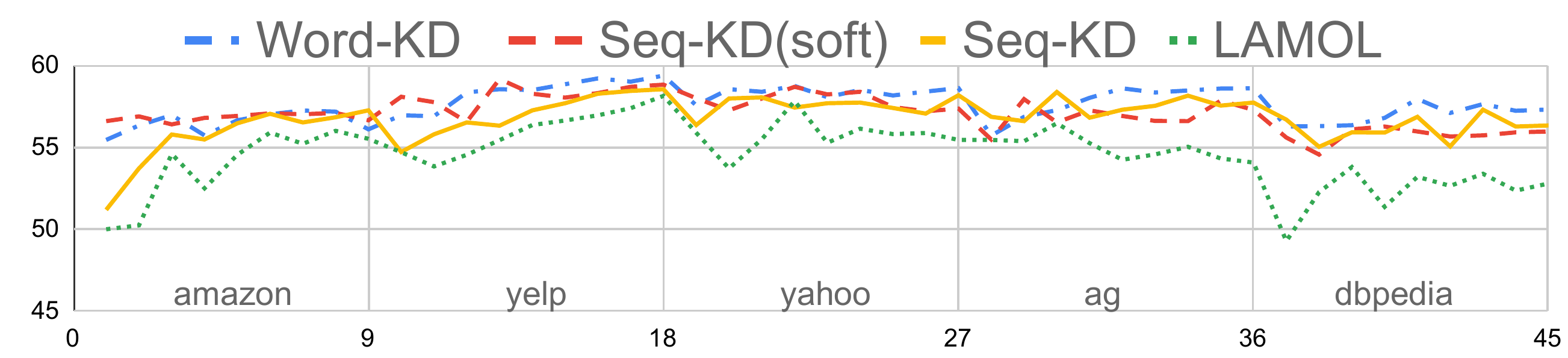}
        \caption{Learning curve of Amazon.}
        \label{fig:curve_tc1}
        \end{subfigure}
    \begin{subfigure}{\columnwidth}
        \includegraphics[width=\linewidth]{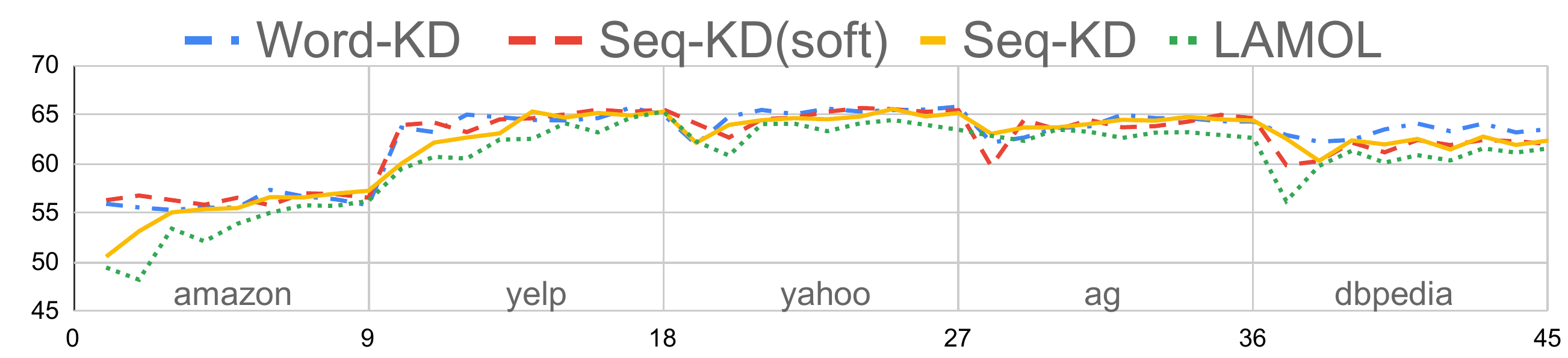}
        \caption{Learning curve of Yelp.}
        \label{fig:curve_tc2}
    \end{subfigure}
    \begin{subfigure}{\columnwidth}
        \includegraphics[width=\linewidth]{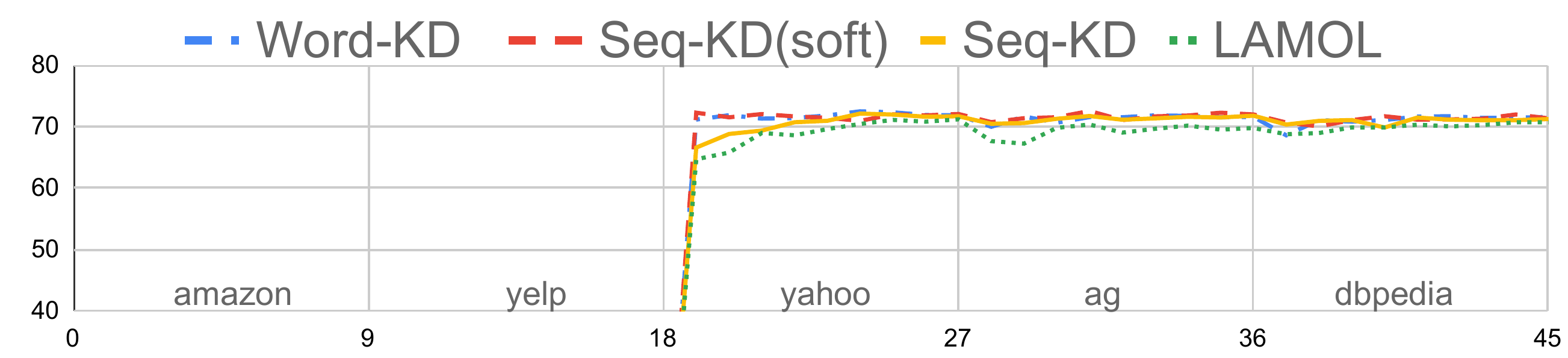}
        \caption{Learning curve of Yahoo.}
        \label{fig:curve_tc3}
    \end{subfigure}
    \begin{subfigure}{\columnwidth}
        \includegraphics[width=\linewidth]{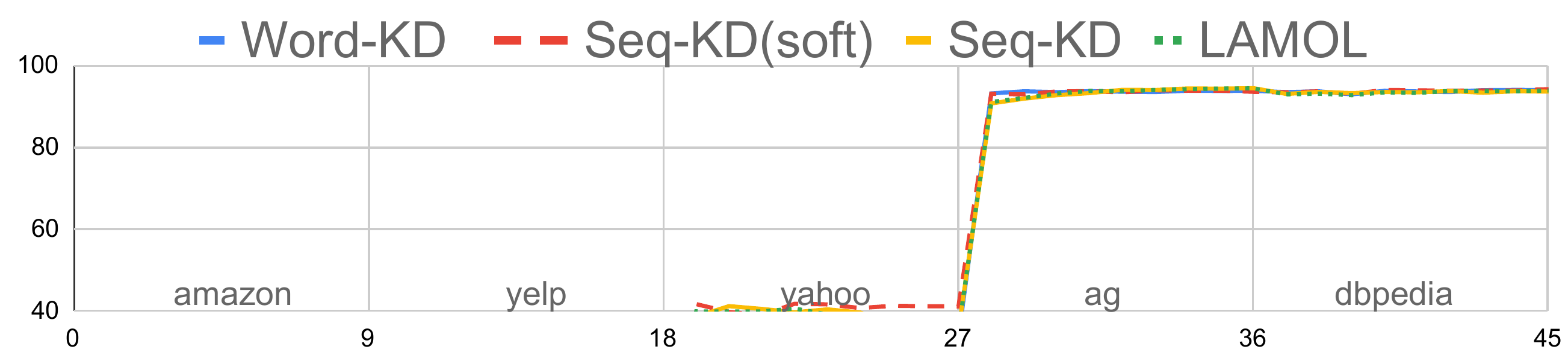}
        \caption{Learning curve of AGNews.}
        \label{fig:curve_tc4}
    \end{subfigure}
    \begin{subfigure}{\columnwidth}
        \includegraphics[width=\linewidth]{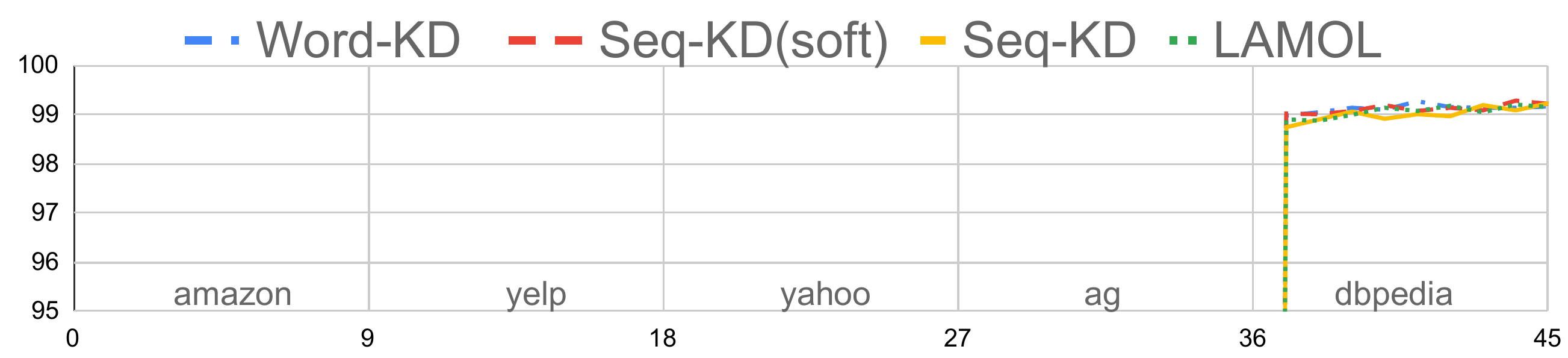}
        \caption{Learning curve of DBpedia.}
        \label{fig:curve_tc5}
    \end{subfigure}
    \caption{The learning curves on the five text classification tasks. X-axis represents epochs, Y-axis represents accuracy.}
    \label{fig:curve_tc}
\end{figure}

The learning curves for five text classification tasks are shown in Figure~\ref{fig:curve_tc}.

\vspace{-15pt}
\section{Detailed Accuracy Analysis for Text Classification Tasks}
\label{appendix:tcanalysis}
The detailed accuracy in group (A) and (B) for five text classification tasks is shown in Table~\ref{tab:tcanalysis_full}.

\begin{table}[t!]
\centering
\small
\setlength\tabcolsep{3pt}
\begin{tabular}{cc}
\toprule
\bf hyperparameter & \bf value \\
\midrule
optimizer & Adam \\
adam epsilon & $1.0 \times 10^{-4}$ \\
learning rate & $6.25 \times 10^{-5}$ \\
training epochs / task & 9 \\
max gradient norm & 1.0 \\
learning rate schedule & warmup linear \\
warmup ratio & 0.005 \\
temperature for KD & 2.0 \\
top-k sampling & k=20 \\
weight decay & 0.01 \\
\bottomrule
\end{tabular}
\caption{The main hyperparameters in the experiment.}
\label{tab:hyper}
\end{table}
\begin{table}[ht!]
\centering
\small
\setlength\tabcolsep{3pt}
\begin{tabular}{lccc}
\toprule
& \bf Acc & \bf Acc in (A) & \bf Acc in (B)\\
\midrule
\bf Amazon \\
\midrule
Teacher & 55.50 & 100.00 & 0.00 \\
\midrule
LAMOL & 52.74 & 66.22 & 35.93 \\
\midrule
+ Word-KD & 57.54 & 73.33 ({\color{green} +7.11}) & 37.85 ({\color{green} +1.92}) \\
+ Seq-KD$_\text{soft}$ & 55.74 & 70.41 ({\color{green} +4.20}) & 37.43 ({\color{green} +1.51}) \\
+ Seq-KD & 56.78 & 71.98 ({\color{green} +5.76}) & 37.82 ({\color{green} +1.89}) \\
\midrule
\bf Yelp \\
\midrule
Teacher & 64.11 & 100.00 & 0.00 \\
\midrule
LAMOL & 61.61 & 75.82 & 36.22 \\
\midrule
+ Word-KD & 63.59 & 79.92 ({\color{green} +4.10}) & 34.43 ({\color{red} -1.80}) \\
+ Seq-KD$_\text{soft}$ & 62.00 & 77.50 ({\color{green} +1.68}) & 34.32 ({\color{red} -1.91}) \\
+ Seq-KD & 62.32 & 77.79 ({\color{green} +1.97}) & 34.68 ({\color{red} -1.54}) \\
\midrule
\bf Yahoo \\
\midrule
Teacher & 71.20 & 100.00 & 0.00 \\
\midrule
LAMOL & 70.29 & 88.28 & 25.81 \\
\midrule
+ Word-KD & 71.28 & 89.63 ({\color{green} +1.35}) & 25.90 ({\color{green} +0.09}) \\
+ Seq-KD$_\text{soft}$ & 71.26 & 89.39 ({\color{green} +1.11}) & 26.45 ({\color{green} +0.64}) \\
+ Seq-KD & 71.13 & 89.52 ({\color{green} +1.24}) & 25.68 ({\color{red} -0.14}) \\
\midrule
\bf AGNews \\
\midrule
Teacher & 93.76 & 100.00 & 0.00 \\
\midrule
LAMOL & 93.63 & 97.15 & 40.70 \\
\midrule
+ Word-KD & 93.91 & 97.67 ({\color{green} +0.52}) & 37.32 ({\color{red} -3.37}) \\
+ Seq-KD$_\text{soft}$ & 93.89 & 97.81 ({\color{green} +0.66}) & 35.00 ({\color{red} -5.69}) \\
+ Seq-KD & 93.45 & 97.15 ({\color{black} +0.00}) & 37.74 ({\color{red} -2.95}) \\
\midrule
\bf DBPedia \\
\midrule
Teacher & 99.11 & 100.00 & 0.00 \\
\midrule
LAMOL & 99.13 & 99.78 & 26.61 \\
\midrule
+ Word-KD & 99.24 & 99.85 ({\color{green} +0.07}) & 31.05 ({\color{green} +4.44}) \\
+ Seq-KD$_\text{soft}$ & 99.18 & 99.85 ({\color{green} +0.07}) & 25.13 ({\color{red} -1.48}) \\
+ Seq-KD & 99.11 & 99.82 ({\color{green} +0.04}) & 19.22 ({\color{red} -7.39}) \\

\bottomrule
\end{tabular}
\caption{The accuracy in the group (A) and (B) detailed in five classification datasets. The teacher scores are from five single-task models.}
\label{tab:tcanalysis_full}
\end{table}

\end{document}